\documentclass[conference]{IEEEtran}
\IEEEoverridecommandlockouts

\usepackage{enumitem}

\usepackage{cite}
\usepackage[numbers]{natbib}
\usepackage{hyperref}
\usepackage{amsmath,amssymb,amsfonts}
\usepackage{algorithmic}
\usepackage{graphicx}
\usepackage{textcomp}
\usepackage{xcolor}
\usepackage{amsthm}
\usepackage{multirow}
\usepackage{tcolorbox}
\usepackage{float}
\usepackage{caption}

\definecolor{grey}{rgb}{0.5,0.5,0.5}
\newtheorem*{definition}{Definition}

\newtheorem*{remark}{Remark}

\def\BibTeX{{\rm B\kern-.05em{\sc i\kern-.025em b}\kern-.08em
    T\kern-.1667em\lower.7ex\hbox{E}\kern-.125emX}}
\begin{document}

\title{Undesirable Memorization in Large Language Models: A Survey
}

\author{
    \IEEEauthorblockN{Ali Satvaty}
    \IEEEauthorblockA{
        \textit{University of Groningen} \\
        Groningen, the Netherlands \\
        a.satvaty@rug.nl
    }
    \and
    \IEEEauthorblockN{Suzan Verberne}
    \IEEEauthorblockA{
        \textit{Leiden University} \\
        Leiden, the Netherlands \\
        s.verberne@liacs.leidenuniv.nl
    }
    \and
    \IEEEauthorblockN{Fatih Turkmen}
    \IEEEauthorblockA{
        \textit{University of Groningen} \\
        Groningen, the Netherlands \\
        f.turkmen@rug.nl
    }
}

\maketitle

\begin{abstract}

While recent research increasingly showcases the remarkable capabilities of Large Language Models (LLMs), it is equally crucial to examine their associated risks. Among these, privacy and security vulnerabilities are particularly concerning, posing significant ethical and legal challenges. At the heart of these vulnerabilities lies memorization, which refers to a model's tendency to store and reproduce phrases from its training data. This phenomenon has been shown to be a fundamental source of various privacy and security issues related to LLMs.

In this paper, we provide a taxonomy of the literature on LLM memorization, structured around three dimensions: \textit{granularity}, \textit{retrievability}, and \textit{desirability}. Building on this framework, we first examine the metrics and methods used to quantify memorization, highlighting how different approaches capture distinct aspects of the phenomenon. This foundation then supports an analysis of the factors that drive memorization and the strategies developed to mitigate its undesirable effects. Finally, we identify promising directions for future research, including methods to balance privacy and performance, as well as the study of memorization in specialized LLM contexts such as conversational agents, retrieval-augmented generation, and diffusion language models.

Given the rapid pace of research in this field, we also maintain a dedicated repository of the references discussed in this survey\footnote{\url{https://github.com/alistvt/undesirable-llm-memorization}}, which will be regularly updated to reflect the latest developments.


\end{abstract}


\section{Introduction}
In recent years, large language models (LLMs) have demonstrated remarkable advancements, driven by the scaling of model parameters, large amounts of data, and extensive training paradigms \cite{gpt3,llama2,chatgpt,rlhf2022}. State-of-the-art models have exhibited capabilities across a broad spectrum of natural language processing (NLP) tasks, consistently pushing the envelope in areas such as text generation, code synthesis, machine translation, question answering, and summarization \cite{wei2022emergent,xu2022systematiccode}. These models are trained on massive datasets, enabling them to perform competitively or even surpass human-level performance in specific tasks \cite{guo2023evaluating,Chang2023A}.

Despite these impressive advancements of LLMs, researchers have shown that there are certain problems  with these models, including hallucination~\cite{huang2023survey}, bias~\cite{tjuatja2024llms}, and privacy and security vulnerabilities~\cite{yao2024survey}. In the context of data privacy, memorization is one of the core sources of concern. Memorization in LLM refers to the model's tendency to store and reproduce exact phrases or passages from the training data rather than generating novel or generalized outputs. While memorization can be advantageous in knowledge-intensive benchmarks, such as factual recall tasks or domain-specific question answering \cite{petroni-etal-2019-language}, it also introduces ethical and legal challenges: models may inadvertently reveal sensitive or private information included in their training data, posing significant privacy and security risks \cite{carlini2021extracting,privacy_under_attack}. In addition, the ability of LLMs to repeat verbatim copyrighted or proprietary text from their training data is closely related to memorization due to which many issues related to intellectual property infringement arise\cite{stochastic_parrots,henderson2023foundation}.

These challenges motivate the need to further explore memorization as a phenomenon in LLMs to effectively tackle the associated challenges. In this paper, we provide an overview of aspects related to memorization, emphasizing the need for further exploration of this topic.

\subsection{Related surveys}
Before recent advances in LLMs, memorization has been explored extensively as a topic in machine learning and deep learning mostly with a focus on security and privacy. \citet{ml_memorization} explore memorization in machine learning. They propose a framework to quantify the influence of individual data samples and detect memorization in various learning settings. \citet{dl_memorization} provides a systematic framework for understanding memorization in deep neural networks, discussing LLM memorization from the view of deep neural networks. Survey papers on the privacy and safety of LLMs often address memorization as a core phenomenon, framing it as both a privacy issue and a foundational factor that supports other non-security/privacy-related challenges they explore \cite{neel2023privacy,smith2023identifying}.  \citet{hartmann2023sokmemorizationgeneralpurposelarge} provide an overview of memorization in general-purpose LLMs. They aggregate memorization-related topics from the copyright, privacy, security, and model performance perspectives. In our survey, we not only incorporate more recent work but also specifically focus on memorization as an undesirable phenomenon, examining it with the aspects of \textit{granularity}, \textit{retrievability}, and \textit{desirability}. Our work is not only relevant for privacy but also for (un)safety and bias as undesirable properties of LLMs when memorization takes place. In addition, we provide an extensive and concrete research agenda for the near future. Table~\ref{tab:existing-surveys} provides a comparison of the scope of our survey with that of previous surveys.

\begin{table*}[t]
\centering
\begin{tabular}{p{3cm} p{3.5cm} p{5cm}}
\hline
\textbf{Paper} & \textbf{Focus} & \textbf{Differences With our Study} \\
\hline
\citet{ml_memorization} &  Memorization in machine learning & We focus on \textit{LLMs}\\
\hline
\citet{dl_memorization} & Memorization in deep neural networks & We focus on \textit{LLMs}\\
\hline
\citet{neel2023privacy} & Privacy Problems in LLMs & \textit{Memorization} is treated as one of the sources of privacy-related problems rather than being the focus. \\
\hline
\citet{smith2023identifying} & Privacy Problems in LLMs & \textit{Memorization} is treated as one of the sources of privacy-related problems rather than being the focus. \\
\hline\citet{hartmann2023sokmemorizationgeneralpurposelarge} & Memorization in general-purpose LLMs & We focus on \textit{undesirable} \textit{memorization} in LLMs, and propose concrete directions for future work. \\
\hline
\end{tabular}
\caption{List of existing surveys vs. our work}
\label{tab:existing-surveys}
\end{table*}

\subsection{Data selection}
The selection of works included in this survey was guided by the goal of capturing a comprehensive view of the literature surrounding memorization in LLMs, with a particular focus on its \textit{undesirable} aspects. However, to provide a well-rounded taxonomy and to contextualize these challenges, we also include works that examine memorization in its desirable forms, such as generalization and knowledge retention, as they offer complementary insights into the mechanisms at play.

\paragraph*{Selection process} The process of identifying relevant papers was iterative, beginning with the most cited early studies by recognized researchers in the field (14 papers). These foundational and influential works were selected as they represent key milestones that have shaped current understandings of memorization in LLMs. Then we included the papers drawn from keyword searches in article repositories. We used \textit{Google Scholar}\footnote{https://https://scholar.google.com/} and \textit{arXiv}\footnote{https://arxiv.org/}, as they are predominantly used by researchers in the computer science 
and computational linguistics domains \cite{sutton2017popularityarxivorgcomputerscience}. For \textit{Google Scholar} we restricted our keyword search to the papers that include the term ``Memorization'' in their title and include ``Language Model'' in their body (memorization is also used as a term in biology). For the scanning of \textit{arXiv}, we collected the papers that included both of the words ``Memorization'' and ``Privacy'' in their abstract and ``language model'' in their body. For inclusion, we selected articles published before \textit{December 2025}. 

Additionally, given the broad scope of the topic, our survey draws from adjacent areas, including studies on data extraction, membership inference, and other forms of data leakage, which intersect with the broader concept of memorization. These works were identified through reference chaining, utilizing bibliographies of key papers, and consulting sections of existing survey articles. We believe this approach allows the survey to address the complexities and nuances of memorization in LLMs.

\begin{figure*}[hbtp]
    \centering
    \includegraphics[width=0.8\textwidth]{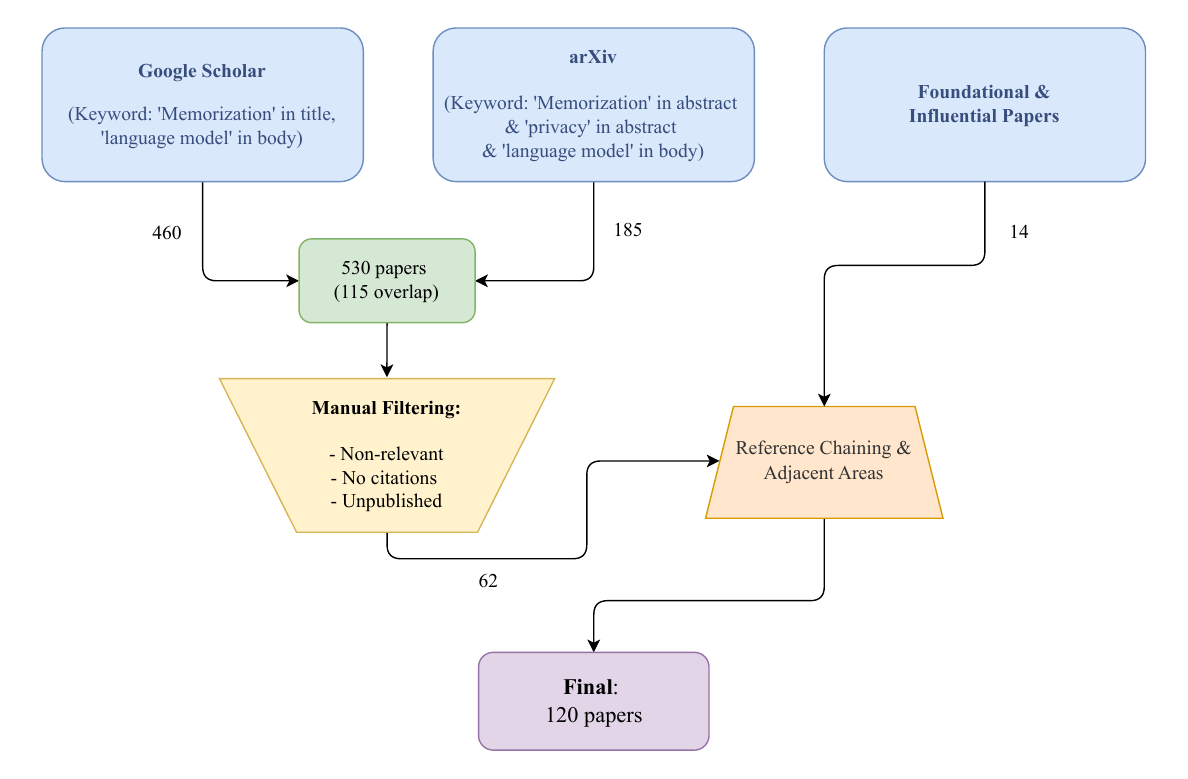}
    \caption{Our data selection process: Starting with 14 foundational papers as the seed, our literature search expanded through keyword search in Google Scholar (460 papers) and arXiv (185 papers), followed by manual filtering, reference chaining, and exploration of adjacent areas, concluding in a final selection of 120 papers.}
    \label{fig:data-selection}
\end{figure*}

\paragraph*{Rejection criteria}
After obtaining the initial repository of papers, we manually iterated through the collection and removed non-relevant papers (e.g., out of scope for ``undesirable memorization", focus on deep learning) and the papers that were neither published at a scientific venue nor had any citations. This process and the obtained statistics are summarized in Figure \ref{fig:data-selection}.


\subsection{Paper organization}
This survey is organized to capture the body of literature around undesirable memorization and offers an extensive view of this phenomenon from different perspectives. Our contribution is driven and structured around the following research questions:

\begin{enumerate}[label=\textbf{RQ\arabic*.}, left=0pt, labelwidth=*, labelsep=1em, align=left]
    \item How is memorization defined in the context of LLMs?
    
    \item  What are the methods used to measure memorization of LLMs?
    
    \item  What are the factors  contributing to memorization?

    \item  What methods are used to prevent or mitigate undesirable memorization?
    
    \item  What are the important aspects of memorization that are still unexplored or require more research?
    
\end{enumerate}

\begin{figure*}
    \centering
    \includegraphics[width=.9\linewidth]{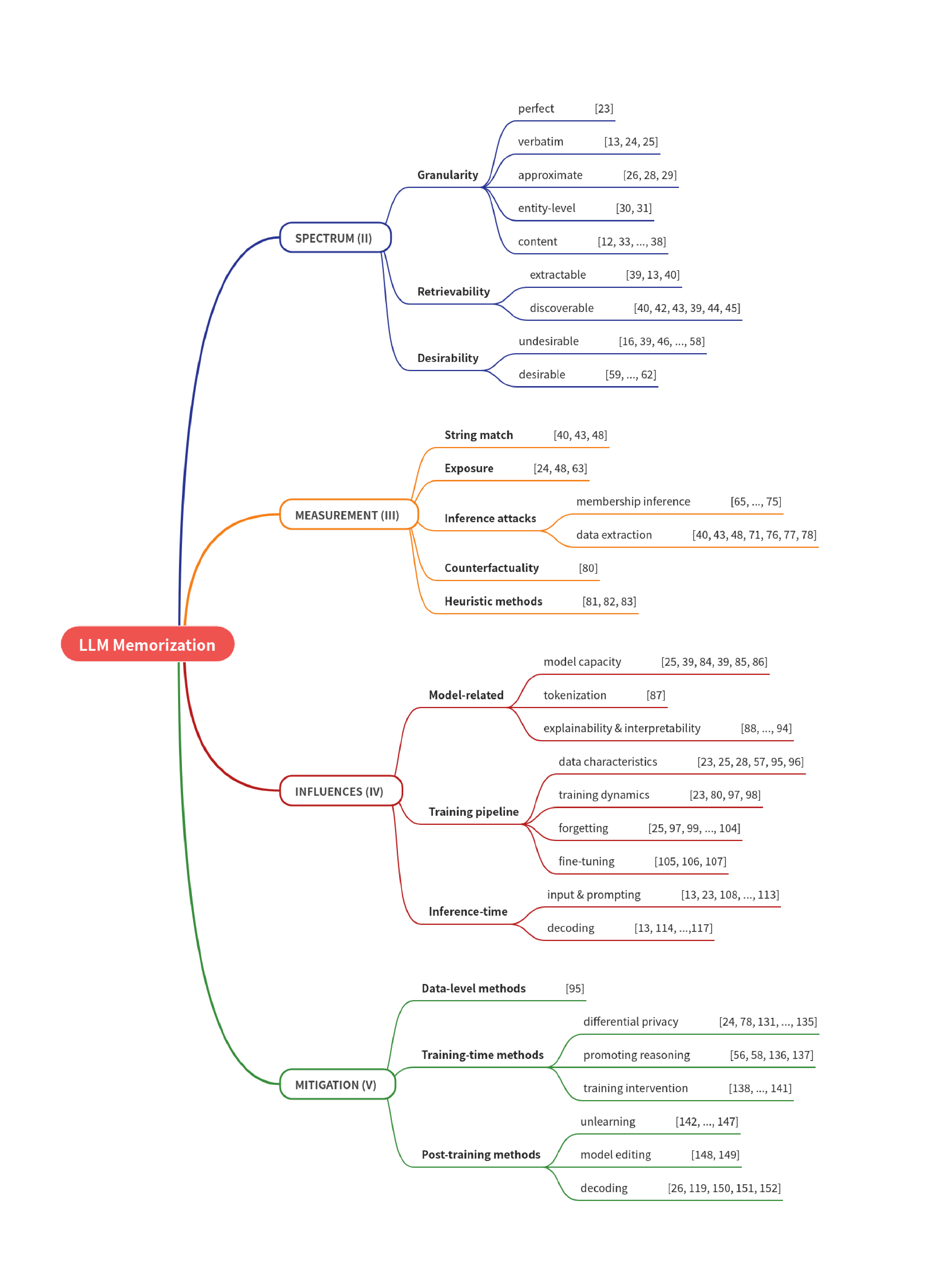}
    \caption{Our categorization of the existing work on LLM memorization}
    \label{fig:overview}
\end{figure*}

In the following sections of this paper, we provide a comprehensive exploration of memorization in LLMs around the mentioned research questions, structured into five main sections. We begin with the \emph{Spectrum of memorization} (section \ref{sec:spectrum}), where we examine the concept from multiple perspectives including \emph{desirability}, \emph{retrievability}, and \emph{granularity}, offering a deep understanding of how memorization occurs. Next, in \emph{Measuring memorization} (section \ref{sec:measuring}), we review various methodologies that have been used in previous studies to quantify and assess memorization within LLMs. We then explore the \emph{Influencing factors and dynamics of memorization} (section \ref{sec:measuring}), identifying key factors and conditions that contribute to this phenomenon. Next, in the \emph{Mitigating memorization} section (section \ref{sec:mitigation}), we discuss strategies and techniques employed to minimize or control memorization in LLMs, addressing concerns related to privacy and generalization. Finally, we conclude with \emph{Future directions and open challenges} (section \ref{sec:future}), outlining potential areas for further research and unresolved questions, before summarizing our findings in the \emph{Conclusion} (section \ref{sec:conclusion}). Figure~\ref{fig:overview} shows a visual overview of our survey scope.

\section{Spectrum of Memorization}

In this section, we provide an overview of the available definitions for memorization in \textit{three} dimensions and categorize the existing work according to these axes. The overall summary of our findings is presented in Table~\ref{tab:dynamics-summary} along with the other aspects.

\label{sec:spectrum}
\subsection{Granularity of memorization} \label{sec:severity}
While memorization in LLMs is widely discussed in the literature, there is no universally accepted definition for it. The definition of memorization can vary based on factors such as the level of detail required, the context in which the information is recalled, and the specific task or application at hand. This section discusses different levels of granularity of \emph{recall} that are explored in the literature evolving around memorization in LLMs.

\paragraph{Perfect memorization}
\emph{Perfect memorization} refers to a setting where a model can generate only from the training data. \citet{k2022deduplicating} define perfect memorization as follows:

\begin{definition} \label{def:perfect-mem}
[\textbf{Perfect memorization}] A model is said to \emph{perfectly memorize} its training data if the generation frequencies of sequences are the same as their appearances in the training data. Sampling outputs from a \emph{perfect memorization} model is identical to sampling from the training data.
\end{definition}

Perfect memorization could be viewed as an upper bound for the level of memorization a language model can exhibit. It acts as an imaginary language model to help compare the extent of memorization in practical scenarios.

\paragraph{Verbatim memorization}
\emph{Verbatim memorization} is defined as a form of memorization that involves the exact reproduction of strings from the training data. It captures instances where the model outputs a training sample without any alterations. This type of memorization is straightforward to identify and is often the primary focus in discussions about memorization in language models~\cite{carlini2019canaries,carlini2021extracting} (Table \ref{tab:dynamics-summary}).

\citet{carlini2021extracting} define verbatim memorization under the term \emph{eidetic memorization} as follows:

\begin{definition} \label{def:Verbatim-mem}
    [\textbf{Verbatim memorization}] Let $LM(p)$ denote the output of a language model when prompted with $p$. A string $s$ from the training set is defined as \emph{verbatim memorized} if there exists a prompt $p$ such that $LM(p) = s$.
\end{definition}
Due to its intuitiveness, various variants of verbatim memorization has been put forward.  \citet{tirumala2022memorization} define another version of verbatim memorization as \emph{Exact memorization} with a focus on the context and sampling method:

\begin{definition} \label{def:exact-mem}
    [\textbf{Exact memorization}] Let $c = (p, s)$ be a context where $p$ is incomplete block that is completed by the sequence $s$. The context $c$ from the training set is considered to be \emph{exactly memorized} by the language model $LM$ if $Argmax(LM(p)) = s$.
\end{definition}

This definition is more strict and does not capture the memorization under different forms of sampling methods. It rather considers only the greedy sampling [see Section \ref{sec:sampling}].

It should be noted that these definitions do not impose any restrictions on the length of the prompt $p$ or the generated output $s$. However, if the generated text is too short, it may not be appropriate to classify it as memorization. Typically, a certain number of tokens is considered when assessing memorization; 50 tokens length prefix and suffix is used predominantly by the researchers (different employed prompt lengths and generation lengths are summarized in table \ref{tab:dynamics-summary}).

Throughout the rest of this paper, we use the word memorization interchangeably with the verbatim memorization (Definition \textit{Verbatim Memorization}), as it is the one that is mostly associated with the undesirable\footnote{Desirability is discussed in section \ref{sec:desirability}} aspects of memorization.

\paragraph{Approximate memorization}
\emph{Approximate memorization} extends beyond verbatim memorization to include instances where the output is similar but not identical to the training data. Verbatim memorization does not capture the subtler forms of memorization, as it is too confined. For example, if two sentences differ by a minor detail, such as punctuation, a misspelled word, or a stylistic variation (e.g., American English vs. British English), these instances would fall outside the strict boundaries of verbatim memorization. However, human judges would likely consider these variations as memorized examples.

\citet{ippolito2023preventing} define \emph{Approximate memorization} when the generation BLEU similarity score  \footnote{a metric used to compare the similarity of texts based on \textit{n-grams} overlaps \cite{papineni-etal-2002-bleu}} with respect to the training data surpasses the 0.75 threshold, which was chosen based on qualitative analyses of the samples. By adopting this definition of memorization, they show that the measurement of memorization can increase by a factor of two compared to only considering verbatim memorization. Similarly, \citet{duan2024uncovering} use token-wise Levenshtein as an approximation distance function to experiment with the dynamics of memorized sequences.

\begin{definition} \label{def:approximate-mem}
    [\textbf{Approximate memorization}] A suffix $s$ for prefix $p$ is labeled as \emph{approximately memorized} if for generation $g = LM(p)$, $Sim(g, s) > \delta $; where $Sim( ,)$ is a textual similarity metric.
\end{definition}

As discussed by \citet{ippolito2023preventing}, this definition can lead to both false positives and false negatives when compared to human judgment, indicating a potential direction for future investigations. Since detecting approximate memorization could be resource intensive, \citet{near-duplicate} introduce a \emph{Mini-hash} algorithm to efficiently detect approximately memorized content by the LLMs based on the Jaccard similarity metric. 


\color{black}

\paragraph{Entity-level memorization}
When considering privacy, the relationships between entities and their connections are often more critical than their exact phrasing in a sentence. This means that even approximate memorization might fail to capture certain privacy risks. For example, suppose the sentence ``John Smith's phone number is 012345678.'' was included in the training set. If an adversary manages to prompt the language model to generate the following sentence: ``I called John Smith yesterday and entered this number into my phone: 012345678,'' the model would still be revealing sensitive information. However, this case would not be detected by the memorization granularities discussed earlier. Despite the lack of verbatim memorization, the model is still violating privacy by exposing the association between John Smith and their phone number. \citet{zhou2023entity} define the \emph{entity-level memorization} as a phenomenon when the model recalls (generates) an entity when prompted with several other entities, when all of these entities have been linked to each other in the training phase.

\begin{definition} \label{def:entity-mem}
    [\textbf{Entity memorization}] Let $s$ be a training sample containing a set of entities $M$. A prompt $p$ can be constructed to include a strict subset of the entities $N \subset M$. Model is said to show \emph{entity memorization} if, when prompted with $p$, outputs a response containing some entities in $M - N$.
\end{definition}

\citet{zhou2023entity} conduct various experiments and show that entity-level memorization happens in different scales affected by model size, context length, and repetitions of the entities in the pretraining data. Similarly, \citet{kim2023propile} target the \emph{personally identifiable information (PII)} leakage in the \textit{OPT} models \cite{zhang2022opt} and show that different types of PII can be extracted if the attackers feed intelligently crafted prompts to the model. They also show that the susceptibility of this leakage is influenced by the type of PII; emails and phone numbers show higher extraction rates compared to addresses.

\paragraph{Content memorization} Beyond the categories described thus far, several broader notions of memorization involve reproducing or inferring general content from the training data, rather than focusing on exact strings (verbatim memorization) or specific entities. This can include, for instance, reproducing factual knowledge in different words than originally presented (often called \emph{factual memorization}), or recalling underlying concepts without necessarily matching the original phrasing (\emph{conceptual memorization} or \emph{knowledge memorization}) \cite{petroni-etal-2019-language, jiang2020know, alkhamissi2022reviewlanguagemodelsknowledge, Luo2023Systematic}.

While these types of memorization can be beneficial for tasks requiring reasoning or knowledge application, they also raise distinct concerns. For example, when a model memorizes factual knowledge or concepts, the source of that information becomes critical \cite{lin2022truthfulqa, min2024silolanguagemodelsisolating}. If the training data contains inaccuracies, biases, or falsehoods, these errors might be presented in the model outputs as well, leading to trustworthiness issues. Additionally, facts or knowledge could be outdated, further compounding the problem. This is particularly problematic in applications where factual accuracy is paramount, such as educational tools or decision-making systems. \citet{bender2021dangers} highlight how language models can sustain misinformation when trained on unverified or noisy datasets, emphasizing the risks tied to the provenance of memorized knowledge.

\subsection{Retrievability} \label{subsec:retrievability}
A common way to analyze memorization in LLMs is through sampling. This involves selecting tokens probabilistically from the model's predicted distribution at each step, to see if a sequence of probabilistically generated tokens can be traced back to the training data. Since the LLM's input and output domains are both discrete, it is important to note that it might not be possible to extract all of the memorized content, given a finite number of generation trials through sampling. 
Based on how an LLM can be prompted to output the training data, memorization can be classified into \textit{extractable} and \textit{discoverable} in terms of retrievability.

\paragraph{Extractable memorization}
\emph{Extractable memorization} refers to the ability to retrieve specific information from a model's training data without direct access to that data. \citet{carlini2023quantifying} defines extractable memorization as follows:

\begin{definition}\label{def:extractable}{\textbf{[Extractable Memorization]}} Let $LM(p)$ denote the output of a language model when prompted with $p$. An example $s$ from the training set $S$ is \emph{extractably memorized} if an \textbf{adversary} (without access to $S$) can construct a prompt $p$ that makes the model produce $s$ (that is, $LM(p) = s$).
\end{definition}

Analyzing extractable memorization usually involves two main challenges: designing prompts that best elicit memorization in a model and verifying if the model output is indeed from the training data.

Research in this area has employed various strategies. \citet{carlini2021extracting} recover training examples from GPT-2 by prompting it with short strings from the public Internet and manually verifying the outputs via Google search. This method confirmed the memorization of about 0.00001\% of GPT-2's training data due to the labor-intensive verification process. \citet{nasr2023scalable} conduct extensive analysis on Pythia, RedPajama, and GPT-Neo models \cite{biderman2023pythia}. They query these models with millions of 5-token blocks from Wikipedia and count for unique 50-grams that the model generates whether they exist in a combined dataset of the models' training data. Their method was more successful, showing that 0.1\% to 1\% of the models' outputs are memorized, with a strong correlation between model size and memorization abilities.

\paragraph{Discoverable memorization}
\emph{Discoverable memorization} measures the extent to which models can reproduce their training data when explicitly prompted with data from their training set. \citet{nasr2023scalable} suggests the following definition for discoverable memorization:

\begin{definition}\label{def:discoverable}{\textbf{[Discoverable memorization]}} Let $LM(p)$ denote the output of a language model when prompted with $p$. For a context $c = (p, s)$ from the training set $S$, we say that $s$ is \emph{discoverably} memorized if $LM(p) = s$.
\end{definition}


 A more specific form of this is \emph{k-discoverable memorization}\footnote{\citet{biderman2023emergent} mention this as \emph{K-extractable}, however, to avoid confusion with the definition of \emph{extractability} (Definition \ref{def:extractable}) we opt to use this terminology for this definition.} which adds a criterion to the number of prefix tokens:

\begin{definition}\label{def:k-discoverable}{\textbf{[k-discoverable memorization]}} For an example $c = (p, s)$ from the training set, string $s$ is said to be \textit{k-discoverable} if $s$ is \textit{discoverable} and $p$ is consisted of $k$ tokens \cite{biderman2023emergent}.
\end{definition}

Considering the limited knowledge of the adversary in the extractable memorization definition, \citet{nasr2023scalable} assume that discoverable memorization provides an upper bound for data extraction. Ideally, discoverable memorization requires querying the model with all of the possible substrings from its entire training set, which is computationally intractable. Also, noteworthy is that discoverable memorization differs from extractable memorization in that the prompt $p$ is known to be from the training set.

\citet{hayes-etal-2025-measuring} introduce the notion of \textit{\textbf{(n,p)-discoverable}} extraction, which characterizes when an adversary can obtain a specific sample by querying the model \textit{n} times with probability at least \textit{p}, under defined sampling schemes. This offers a more operational and practice-oriented framework for assessing extraction risk.

\citet{carlini2023quantifying} investigate the upper bounds of data extraction in GPT-Neo models through discoverable memorization. They find that (a) LLMs discoverably memorize roughly 1\% of their training datasets; (b) there is a log-linear correlation between data extraction and model size, repetition of data, and prefix context length. Other studies on different models (PaLM, MADLAD-400) corroborate the 1\% memorization rate when prompting with about 50 tokens of context \cite{anil2023palm,kudugunta2023madlad400}.

\paragraph{Discoverable and extractable Memorization}
\citet{nasr2023scalable} compare their extractable memorization results with the discoverable memorizations from \citet{carlini2023quantifying} for the GPT-Neo 6B parameter model. This comparison revealed that (1) some sequences are both discoverably and extractably memorized; (2) some sequences are discoverably memorized but not extractably memorized, and vice versa; (3) the overlap between these two types of memorization provides insights into the model's information retention and retrieval mechanisms. This comparison highlights the complementary nature of these two approaches in understanding a model's memorization capabilities and the retrievability of information from its training data.

\begin{remark}
Based on the definitions presented here, one might perceive \emph{extractable/discoverable memorization} and other concepts within the \emph{granularity dimension} (section \ref{sec:severity}) as equivalent. However, it is important to distinguish them. \emph{Extractable} and \emph{discoverable memorization} emphasize the methods used to retrieve memorized samples. In the case of \emph{granularity of memorization}, the retrieval method is irrelevant; the primary concern is what granularity of the data has been memorized, irrespective of how it is produced by the model. Thus, each of the levels of the \emph{granularity}, can be encompass both \emph{extractable} and \emph{discoverable memorization}.
\end{remark}

\subsection{Desirability} \label{sec:desirability}

Although memorization of factual information can be helpful for the model to perform more accurately on benchmark tasks such as question answering, other data might be retained without any clear purpose, and this might cause issues with privacy and copyright. In this regard, memorization could be categorized into desirable and undesirable subcategories. \cite{peris2023privacy,carlini2023quantifying,Huang2022Detecting}.

\paragraph{Undesirable memorization}

Memorization has been demonstrated to be partly an undesirable phenomenon, often resulting in a range of issues, including:
\begin{itemize}
    \item{\textbf{Privacy risks:}} LLMs might inadvertently memorize and potentially reveal sensitive personal information present in the training data \cite{henderson2023foundation}.
    \item {Security vulnerabilities:} Unintended memorization of confidential information like API keys or passwords could lead to security breaches \cite{carlini2019secret,ramaswamy2020trainingproductionlanguagemodels,huang2024codesecretmem}.
    \item {\textbf{Copyright violation:}} Memorization of copyrighted material may lead to legal challenges, especially if the model can reproduce substantial portions of protected works \cite{henderson2023foundation, mueller2024llmsmemorizationqualityspecificity, freeman2024exploring, cooper2025filescomputercopyrightmemorization}.
    \item {\textbf{Bias and fairness:}} Based on the distribution of the data, memorization could introduce bias issues in the model output.
    \item{\textbf{Overperformance on the benchmarks:}} LLMs often memorize benchmark datasets, leading to overfitting and inflated performance on seen data \cite{elangovan2021memorization,bordt2024elephants}.
    \item{\textbf{Performance degradation on specific contexts:}} When models lean on memorized content, they may produce incorrect outputs on tasks that require reasoning, or generate invalid information due to reliance on outdated memorized knowledge \cite{du-etal-2025-reason, prashanth2024recitereconstructrecollectmemorization, hong-etal-2025-reasoning}.
\end{itemize}

\paragraph{Desirable memorization}
\citet{ranaldi-etal-2023-precog} in their experiments show that memorization is could be beneficial for model performance. Moreover, as discussed above, even though unintended, memorization sometimes allows models to store and utilize vast amounts of knowledge, acting as a desirable phenomenon.
Key aspects of desirable memorization include:
\begin{itemize}

\item{\textbf{Transparency and auditing:}} When models retain recognizable traces of their training data, users and auditors can directly test whether specific information was included in the training corpus. This observability enables practical checks on data provenance, supports accountability, and offers a way for individuals to verify whether their own data may have been used, leading to a careful selection of training data by the model owners.
\item{\textbf{Knowledge retention:}} LLMs are often deliberately designed to retain facts, concepts, and general knowledge to enhance their performance in tasks like question answering and information retrieval \cite{Chen2023Beyond, luetal2024scaling}.
\item{\textbf{Language generation:}} Deliberate memorization of linguistic patterns, grammar rules, and vocabulary is crucial for the model's ability to generate coherent and contextually appropriate text.
\item{\textbf{Alignment goals:}} Intentional memorization can be utilized in the AI alignment phase to inject desired behaviors and values into the model \cite{ouyang2022training}.
\end{itemize}

\begin{remark}
    While the distinction between undesirable and desirable memorization is conceptually useful, the practical implications are more nuanced. Many effects of memorization depend on who is evaluating the model (e.g., model developers, end-users, data subjects, copyright holders) and in which context the memorized content appears (e.g., bias mitigation versus alignment, knowledge retention versus privacy). A behavior that benefits general users can simultaneously disadvantage copyright holders: memorization that enables auditing and data-provenance checks for users may expose sensitive information from data subjects; memorization that stabilizes alignment behaviors may also reinforce existing biases; memorization that improves factual recall may increase the risk of generation of false information due to out-datedness. In this sense, even when memorization provides functional advantages, it always carries an undesirable risk.
\end{remark}

\color{black}

\begin{figure}
    \centering
    \includegraphics[width=.8\linewidth]{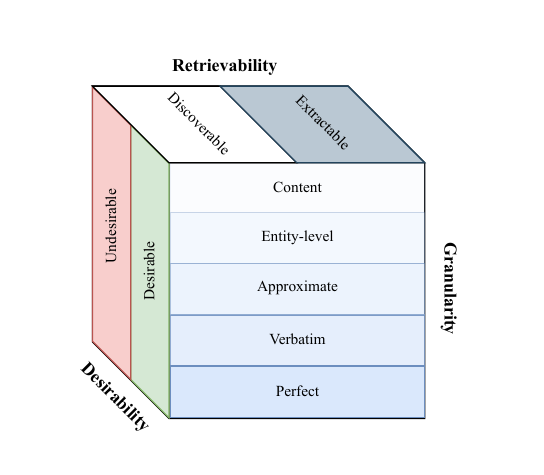}
    \caption{The spectrum of memorization could be viewed as a 3-dimensional cube.}
    \label{fig:spectrum}
\end{figure}

\begin{tcolorbox}[colback=white, title=Answer to RQ1.]
By reviewing the literature, we can define memorization across three dimensions: granularity, retrievability, and desirability. These dimensions are illustrated in Figure \ref{fig:spectrum}.
\end{tcolorbox}

\section{Measuring memorization}
\label{sec:measuring}

Measuring memorization in generative language models was originally performed using the exact match metric \cite{carlini2019secret}. However, as we will discuss in Section \ref{sub:exact-match} this has its own limits and issues, therefore we discuss different methods that can be used to provide an approximation of the memorization of a specific model through so-called white-box attacks. 

\subsection{String match}\label{sub:exact-match}
Measuring verbatim memorization (Section \ref{def:Verbatim-mem}) with the methods discussed in the retrievability section (Section \ref{subsec:retrievability}), requires exhaustively interacting with the model by inputting different prompts and comparing the model output to the training data. Then the attack success rate is measured by dividing the portion of memorized text by the size of the training data. Since there are infinite combinations of tokens that one can feed the model, this method usually falls short in providing the accurate amount of memorization, rather it can provide a good approximation on memorization lower bound \cite{carlini2019secret,nasr2023scalable, hayes-etal-2025-measuring}. As the exact match metric is too sensitive to small perturbations in the generation, approximate match (Section \ref{def:approximate-mem}) has been introduced \cite{ippolito2023preventing} to capture more memorized samples by bypassing negligible changes in the outputs.

\color{black}

\subsection{Exposure metric}
\label{sec:exposure}
Introduced by \citet{carlini2019canaries}, \emph{exposure} provides a quantitative measure of how much a model has memorized specific sequences from its training data. This metric is particularly useful for assessing the memorization of rare or unique information.

Exposure is computed based on the negative log-rank of the generation of a sequence according to a language model. To practically apply the exposure metric, researchers often use the ``canary extraction test'' \cite{carlini2019canaries}. This involves inserting known \textit{secret numbers} as `\textit{canary}' sequences into the training data and then measuring their exposure in the trained model.

\begin{definition}
\label{def:exposure}
    [\textbf{Exposure}] Given a canary $s[r]$, a model with parameters $\theta$, and the randomness space R, the exposure of $s[r]$ is \cite{carlini2019secret}:
     \begin{equation} 
    exposure_{\theta}(s[r]) = log_2 |R| - log_2 
 rank_{\theta}(s[r]) 
    \end{equation} 
\end{definition}

\citet{d_exposure} propose the so-called ``d-exposure'' metric as a measure of memorization for discriminative tasks such as text classification, since in those situations we don't have access to the perplexity of a given text. 

\subsection{Inference attacks}
Inference attacks are another approach to measuring memorization, focusing on the model's ability to reveal information about its training data. These attacks typically perform under the assumption that when a model displays high confidence in its outputs, those outputs are likely to be from the training data, which means the model has memorized them. These attacks can be categorized into two main types:

\paragraph{Membership Inference Attacks (MIA)} These attacks aim to determine whether a specific data point was part of the model's training set. \citet{shokri2017membership} introduced this concept for machine learning models, and it has since been adapted for language models. Doing MIA on LLMs typically involves determining the model's confidence in a given text and using it as an indicator of whether the text was part of the training data. This is often done using the log-likelihood of the sequence tokens \cite{mattern2023membership,shachor2024improvedmembershipinferenceattacks,shejwalkar2021membership,jagannatha2021mia,analyzing,Auditing2019,wang2024pandora, wei2025hubblemodelsuiteadvance, xie-etal-2024-recall, satvaty2025elmiaquantifyingmembershipinference, zhang2025mink}.

\paragraph{Extraction Attacks} These attacks attempt to extract specific pieces of information from the model that were present in its training data. Different works have demonstrated the feasibility of such attacks on language models, showing that private information could be extracted through prompting with different strategies \cite{ishihara-2023-training,nasr2023scalable,Parikh2022Canary,carlini2019secret,lukas2023analyzing,wang2024pandora, hayes-etal-2025-measuring}.


The effectiveness of inference attacks can serve as a proxy measure for memorization. Models that are more susceptible to these attacks are generally considered to have higher levels of memorization.

\subsection{Counterfactuality} \label{sec:counterfactual}
Previous work analyzed the memorization of large language models on sensitive information (e.g. phone numbers) in the training data \cite{carlini2021extracting} or synthetically injected `canaries' \cite{carlini2019canaries,henderson2018ethical}. However, not all the memorized texts are equally interesting. \citet{zhang2023counterfactual} propose another measure of memorization which is \textit{counterfactual memorization}. The idea is to see how the presence or absence of a sample of the dataset, affects the performance of the model on the same sample. This measure has similarities with the definition of differential privacy.\footnote{Differential privacy as a mitigation strategy will be discussed in Section~\ref{sec:dp}.}

In their experiments, the authors create different subsets of a bigger dataset and then they fine-tune the LM on each of these. Then they consider an item $x$ (e.g. a document from Wikipedia as the dataset) from the datasets and based on the presence or absence of $x$ in the subsets, they divide the subsets into two groups: IN and OUT. Then they test and report the performance on the IN and OUT group of models by averaging. Their experiments on 400 trained models show the counterfactually memorized data, are generally unconventional texts such as all-caps, structured formats (i.e. tables or bullet lists), and multilingual texts. In conclusion, counterfactuality could be used as a metric for measuring memorization, however, the interplay between this metric and the different types of memorization is unexplored and worth more research in the future.

\subsection{Heuristic methods} \label{sec:promptcompression}
\citet{schwarzschild2024rethinking} introduce the Adversarial Compression Ratio (ACR) as a novel metric to assess memorization in large language models (LLMs). ACR evaluates whether a string from the training data is memorized by determining if it can be elicited with a significantly shorter adversarial prompt. The method uses the GCG algorithm \cite{zou2023universaltransferableadversarialattacks} to find the most compressed prompt. This approach aligns conceptually with Kolmogorov complexity\footnote{Kolmogorov complexity measures the shortest possible set of instructions (or program) needed for a computer (\emph{Turing machine}) to generate a specific output.}, as it measures the minimum description length of a string but adapts the concept for practical use in LLMs by focusing on adversarial prompts. Similarly, \citet{feng2025evaluating} utilize soft token sparsity as a measure of memorization, operating on the premise that LLMs generate memorized content with higher confidence and determinism.


\begin{tcolorbox}[colback=white, title=Answer to RQ2.]
The most dominant method used to measure memorization of LLMs is the \emph{string match}, however, inference attacks, exposure metrics, counterfactuality, and prompt compression methods can also be used to measure memorization in a more informative way. 
\end{tcolorbox}

\section{Influencing factors and dynamics of memorization}
\label{sec:influence}
Understanding the factors that influence memorization in LLMs is a crucial to overcome the undesirabilities of this phenomenon. This section explores various aspects that affect memorization, organized into \textit{model-related aspects, training pipeline influences}, and \textit{inference-time causes}.

\subsection{Model-related factors \& dynamics}

\subsubsection{Model capacity}\label{sec:model-size}
The first significant factor influencing memorization is model size. \citet{carlini2023quantifying, tirumala2022memorization, kiyomaru-etal-2024-comprehensive-analysis}, demonstrated that larger models are more prone to memorization and do so more rapidly. This trend persists across different architectures and datasets. \citet{carlini2023quantifying} find that the relationship between model size and memorization grows consistently on a log-linear scale, with larger models memorizing a greater portion of the data.

\citet{tirumala2022memorization} further highlight that larger models not only memorize more but do so faster in the training process. Interestingly, while memorization increases with model size, it does not necessarily correlate with improved performance. This was shown by \citet{carlini2023quantifying}  by comparison of models with similar capacities but differing performance levels because of their architectures.

In the same line, several works analyze the memorization capacity of Transformers from a theoretical perspective. \citet{mahdavi2024memorization} explore the memorization abilities of multi-head attention mechanisms, showing that the number of memorized sequences scales with the number of heads and context size under specific assumptions about input data. Similarly, \citet{kim2023provable} quantify the theoretical lower bound of memorization capacity of Transformers on sequence-to-sequence mappings and examine this capacity across classification and language modeling tasks with empirical validation.

Overall, the findings suggest that the ability of LLMs to memorize is strongly linked to their size, potentially due to the high capacity of these models to store detailed information from training data.




\subsubsection{Tokenization}
\citet{kharitonov202BPE} explore the impact of the tokenizer on memorization. They experiment with the size of the sub-word vocabulary learned through Byte-Pair Encoding (BPE) and demonstrate that increasing the sub-word vocabulary significantly affects the model's ability and inclination to memorize training data. Furthermore, models with larger vocabularies are more likely to reproduce training data when given specific prompts. The authors suggest that this stems from the reduction in sequence lengths as BPE vocabulary size increases.

\color{black}

\subsubsection{Explainability and interpretability}
A growing body of interpretability work aims to uncover how LLMs memorization is represented, triggered, and propagated. \citet{huang-etal-2024-demystifying} study verbatim memorization in LLMs using controlled pre-training with injected sequences. They find that memorization requires significant repetition, increases in later checkpoints, and is tied to distributed model states and general language modeling capabilities. Their stress tests show unlearning methods often fail to remove memorized information without degrading model performance, highlighting the difficulty of isolating memorization. 
\citet{haviv-etal-2023-understanding} propose a framework to probe how memorized sequences are recalled in transformers, showing that memory recall follows a two-step process: early layers promote the correct token in the output distribution, while upper layers amplify confidence. They find that memorized information is primarily stored and retrieved in early layers. Building on this layer-wise perspective, \citet{dankers-titov-2024-generalisation} investigate where memorization occurs across model layers, demonstrating that memorization is a gradual and task-dependent process rather than localized to specific layers. Using centroid analysis, they show that deeper layers contribute more to memorization when models generalize well to new data. Complementing these findings, \citet{stoehr2024localizingparagraphmemorizationlanguage} show that memorization in LLMs, while distributed across layers, is driven by distinct gradients in lower layers and influenced by a low-layer attention head focusing on rare tokens. Perturbation analysis reveals that distinctive tokens in a prefix can corrupt entire continuations, and memorized sequences are harder to unlearn and more robust to corruption than non-memorized ones. 

Beyond layer-level analysis, other work explores how memorization manifests in the model's internal representations. \citet{chen-etal-2024-multi-perspective} identify clustering of sentences with different memorization scores in the embedding space and observe an inverse boundary effect in entropy distributions for memorized and unmemorized sequences. They also demonstrate that hidden states of LLMs can be used to predict unmemorized tokens, shedding light on the dynamics of memorization within the model. 

More mechanistic perspectives further refine this picture. \citet{huang-etal-2025-neuron} show that memorization and generalization in LLMs are encoded in largely distinct subsets of neurons. By intervening on these neurons at inference, they demonstrate that a model's tendency to memorize or generalize can be selectively amplified or suppressed without substantially affecting overall performance.
\citet{lasy2025understandingverbatimmemorizationllms} show that verbatim memorization in LLMs is driven by specific neural circuits, with distinct subgraphs responsible for initiating versus sustaining memorized sequences, revealing mechanistic pathways underlying exact data recall.

Together, these findings offer ongoing insights into the mechanisms of memorization, enhancing interpretability and guiding future research.

\subsection{Training pipeline}

\subsubsection{Training data characteristics}\label{sec:training-data}
The nature of the training data heavily influences memorization. \citet{lee2022deduplicating} develop tools to deduplicate training data and show that models trained on deduplicated data would produce memorized text ten times less frequently. \citet{k2022deduplicating} show that a sequence appearing 10 times in the training data is, on average, generated approximately 1000 times more frequently than a sequence that appears only once. These findings emphasize the central role of duplication in driving memorization.

A study by \citet{tirumala2022memorization} on the memorization of different parts of speech in a text reveals that nouns and numbers are memorized significantly faster than other parts of speech, likely because they serve as unique identifiers for specific samples. \citet{prashanth2024recitereconstructrecollectmemorization} categorize memorization into recitation (memorization of highly duplicated sequences), reconstruction (generation using learned templates or patterns), and recollection (memorization of rare, non-template sequences). They argue that memorization is a multi-faceted phenomenon influenced by duplication, template patterns, and rarity, requiring nuanced analysis.

As simple sequences, such as repeated patterns or numbers, are easily memorized by models but often lack substantive content or sensitive information, distinguishing the memorization of these trivial sequences from more complex and meaningful ones would be essential. \citet{duan2024uncovering} observed that data with lower \emph{z-complexity}\footnote{Z-complexity is a way to measure how much temporary memory (workspace) a computer needs to generate a specific output using the shortest possible instructions. Unlike \emph{Kolmogorov complexity}, which only looks at the length of the instructions, Z-complexity also considers the memory used during the process.} leads to faster decreases in training loss, as more compressible patterns are memorized more quickly. They further demonstrated that strings of varying complexity exhibit distinct memorization curves, with lower-complexity strings being memorized more easily even for smaller repeats, following a log-linear relationship in memorization probability. This is indirectly supported by the study of \citet{arnold-2025-memorization} on the memorization as a function of ``Intrinsic Dimension''\footnote{Intrinsic dimension is the minimum number of variables (or degrees of freedom) needed to describe the essential structure of a dataset or a model’s representations.}, observing a negative correlation for the low-duplicated data points group.

Across these perspectives, the evidence converges: duplication, content distinctiveness, and structural simplicity jointly determine how easily different data types are memorized.

\color{black}
\subsubsection{Training process dynamics}
Beyond data properties, the temporal dynamics of training also plays a critical role in shaping when and how memorization emerges. \citet{zhang2023counterfactual,k2022deduplicating} show that memorization grows consistently with the number of training epochs, which makes sense, as more epochs push the model to potential overfitting. \citet{jagielski2023measuring} demonstrate that examples seen during the earlier stages of training are less prone to memorization and rather they are forgotten over time. These findings indicate that memorization increases with more training, while early-seen examples being more likely to be forgotten. \citet{TrainingPhaseMem2024} show higher memorization rates happens early and late in training, with lower rates mid-training. Therefore, suggesting that placing sensitive data in the middle stages of training could reduce its vulnerability to extraction attacks.

\subsubsection{Forgetting mechanisms} \label{sec:forgetting}
In machine learning, forgetting mechanisms are the processes through which models lose or discard previously learned information~\cite{de2021continual}. These mechanisms can occur unintentionally as part of the natural training dynamics or be purposefully induced to meet specific objectives, such as improving model generalization or addressing privacy concerns~\citep{timm2018intentional,beierle2019intentional,nguyen2022survey}. \citet{blanco2024digital} provide a recent, detailed overview of forgetting in LLMs.

\citet{kirkpatrick2017overcoming} initially introduced ``catastrophic forgetting'' in the context of neural networks and continual learning. They propose a method to protect important model weights to retain knowledge. This approach has been effective in maintaining performance on older tasks, even after long periods of non-use. While this early work focused on task retention, later studies explore forgetting within pretraining settings.

\citet{tirumala2022memorization} observe the forgetting mechanisms of a special batch through the learning process and show that it follows an exponential degradation, reaching a constant value baseline. They show that the mentioned baseline scales with the model size. These results suggest that forgetting is structured rather than arbitrary, and that larger models may retain certain information more persistently (section \ref{sec:model-size}).

\citet{jagielski2023measuring} address the dual phenomena of memorization and forgetting in LLMs through stronger privacy attacks and several strategies for measuring the worst-case forgetting of the training examples. The study introduces a method to measure to what extent models forget specific training data, highlighting that standard image, speech, and language models do indeed forget examples over time, though non-convex models might retain data indefinitely in the worst case. The findings suggest that examples from early training phases, such as those used in pre-training large models, might enjoy privacy benefits but could disadvantage examples encountered later in training. 

Overall, these studies illustrate that forgetting is shaped by both the training timeline and the model's internal mechanisms, and can sometimes be leveraged to balance memorization, generalization, and privacy.

\subsubsection{Fine-tuning and transfer learning}
\citet{mireshghallah2022empirical} evaluate how different fine-tuning methods—full model, model head, and adapter fine-tuning—vary in terms of memorization and vulnerability to privacy attacks. Their research, using membership inference and extraction attacks, finds that head fine-tuning is most susceptible to attacks, whereas adapter fine-tuning is less prone. \citet{zeng2024exploring} conduct a comprehensive analysis of fine-tuning T5 models \cite{raffel2020T5} across various tasks, including summarization, dialogue, question answering, and machine translation, finding that fine-tuned memorization varies significantly depending on the task. Additionally, they identify a strong link between attention score distributions and memorization, and propose that multi-task fine-tuning can mitigate memorization risks more effectively than single-task fine-tuning.

\subsection{Inference-time factors}

\subsubsection{Input and prompting strategies}
Prompting methods can strongly influence the generation of memorized content at the inference stage. \citet{carlini2021extracting, mccoy2023much, k2022deduplicating} show that longer prompts increase the likelihood of triggering memorized sequences, making it easier for language models to regurgitate training data. Moreover, methods like prefix tuning \cite{li2021prefix} and prompt engineering have been employed to maximize memorization. \citet{ozdayi2023controlling} introduce a novel approach using prompt tuning to control memorized content extraction rates in LLMs. \citet{DynamicSoftPrompt2024} introduce a dynamic, prefix-dependent soft prompt approach to elicit more memorization. By generating soft prompts based on input variations, this method outperforms previous techniques in extracting memorized content across both text and code generation tasks. \citet{kassem2024alpacavicunausingllms} investigate how instruction-tuning influences memorization in LLMs, showing that instruction-tuned models can reveal as much or more training data as base models. They propose a black-box prompt optimization method where an attacker LLM generates instruction-based prompts, achieving higher memorization levels than direct prompting approaches. \citet{weller2024according} propose `according-to' prompting, a technique that directs LLMs to ground responses in previously observed text.

\subsubsection{Decoding methods} \label{sec:sampling}
While memorization phenomenon stems from the model internals, decoding methods have an important role in arousing it. In their experiments, \citet{carlini2021extracting} initially opt for \emph{greedy decoding} to maximize the regeneration of training data. One limitation is that this decoding scheme generates low-diversity outputs; thus, they also experiment with the \emph{decaying temperature} and \emph{Top-n} decoding methods, the latter of which shows to be more successful. \citet{yu2023bag} experiment with different decoding schemes, including \emph{decaying temperature}, \emph{top-n}, \emph{nucleus-$\eta$}, and \emph{typical-$\phi$} decoding \cite{fan2018hierarchical, holtzman2019curious, meister2023locally} and use an auto-tuning method on these to find the optimal decoding method that yields to the maximization of training data reproduction. As could be drawn from table \ref{tab:dynamics-summary}, most of the previous works use greedy decoding to heighten memorized content generation, suggesting it could serve as a standard approach for future researchers as well.

\begin{table*}[htbp]
\centering
\begin{tabular}{p{3.4cm} p{5.2cm} p{2.8cm}}
\hline
\multicolumn{1}{l}{Factor from Section~\ref{sec:influence}} & \multicolumn{1}{l}{Key findings}                                                                                                           & \multicolumn{1}{l}{Representative Works} \\ \hline
Model capacity             & Larger models memorize more.                                                                                                                & \cite{carlini2023quantifying,tirumala2022memorization} 
\\
 Training data              & Duplicated data amplifies memorization.                                                                                                     &
\cite{lee2022deduplicating,k2022deduplicating,tirumala2022memorization} 
\\
Input and prompting        & Longer prompts and prompt tuning can facilitate recall of the memorized suffix.                                                            & 
\cite{carlini2021extracting, mccoy2023much, k2022deduplicating,li2021prefix,ozdayi2023controlling,weller2024according}
                            \\
Tokenization               & Bigger tokenizer vocabulary leads to more memorization.                                                                                     & 
\cite{kharitonov202BPE} \\
Decoding methods           & Greedy decoding is dominantly employed to extract memorized data. & 
\cite{carlini2021extracting,yu2023bag} 
\\
Fine-tuning                & The amount of memorization after fine-tuning significantly varies depending on the task.                                                                        & 
\cite{mireshghallah2022quantifying,zeng2024exploring} 
\\
Training process           & Earlier phases of training are less prone to memorization.                                                                                    & 
\cite{zhang2023counterfactual,k2022deduplicating,jagielski2023measuring} \\
Forgetting mechanisms      & Forgetting follows an exponentially decaying curve.                                                                                         & 
\cite{tirumala2022memorization,jagielski2023measuring}
\\                       \hline
\end{tabular}
\caption{Factors of influence and dynamics of memorization and their key findings (discussed in Section~\ref{sec:influence})}
\label{tab:findings}
\end{table*}

\thispagestyle{empty}
{
\captionsetup{width=\textwidth} 
\begin{table*}[htbp]
\centering
\footnotesize
\begin{tabular}{p{2.5cm}|p{1cm}p{1cm}p{2cm}p{2cm}p{1.0cm}p{1.0cm}p{1cm}}
\hline
\\
\multicolumn{1}{l|}{Work}                                                    & \multicolumn{1}{l}{Retrievability} & \multicolumn{1}{l}{Granularity} & Model                                    & Dataset                                             & Decoding                         & Prompt Len        & Match Len                \\ \hline \\
\citet{carlini2023quantifying}                              & discoverable                       & verbatim                        & GPT-Neo                                  & PILE                                                &                                  & {50 - 450}      & 50                       \\
\citet{tirumala2022memorization}                            & discoverable                       & verbatim                        & roberta                                  & wikitext-103                                        & greedy                           &                   &                          \\
\citet{NucleusMem2024}                                      & discoverable                       & verbatim                        & GPT-Neo                                  & OpenMemText \cite{NucleusMem2024} & nucleus                          & 50                & {50 - 450}             \\
\citet{zhou2023entity}                                      & extractable                        & entity                & GPT-Neo , GPT-J                          & PILE                                                &                                  &                   &                          \\
\citet{DynamicSoftPrompt2024}                               & discoverable                       & verbatim                        & GPT-Neo, GPT-J, Pythia, StarCoderBase & PILE, the-stacksmol                                & greedy                           &                   &                          \\
\citet{kassem2024alpacavicunausingllms}                     & extractable                        & approximate                     & Alpaca, Vicuna, Tulu                   & RedPajama, RefinedWeb, Dolma                      &                                  & {66 , 100, 166} & {133, 200, 366}        \\
\citet{TrainingPhaseMem2024}                                & discoverable                       & verbatim                        & OLMo                                     & Dolma                                               &                                  & 32                & 32                       \\
\citet{duan2024uncovering}                                  & discoverable                       & approximate                     & Pythia-1b, Amber-7b                     & PILE                                                &                                  & 32                & 64                       \\
\citet{kiyomaru-etal-2024-comprehensive-analysis}           & discoverable                       & approximate, verbatim          & Pythia, LLM-jp                          & PILE, Japanese Wikipedia                           & greedy                           & {100 - 1000}    & 50                       \\
\citet{stoehr2024localizingparagraphmemorizationlanguage}   & discoverable                       & verbatim                        & GPT-Neo-125M                             & PILE                                                & greedy                           & 50                & 50                       \\
\citet{chen-etal-2024-multi-perspective}                    & discoverable                       & verbatim                        & Pythia[dedup]                             & PILE                                                & greedy                           & {32,48,64,96}   & {32,48,64,96}          \\
\citet{huang-etal-2024-demystifying}                        & discoverable                       & verbatim                        & Pythia[dedup]                             & PILE                                                & greedy                           & {8, 16, 32, 64} & 32                       \\
\citet{prashanth2024recitereconstructrecollectmemorization} & discoverable                       & verbatim                        & Pythia[dedup]                            & Memorized set of PILE               & greedy                           & 32                & 32                       \\
\citet{carlini2021extracting}                               & extractable                        & verbatim                        & GPT-2                                    & model generations                                   & greedy, temperature                  &                   & 256                      \\
\citet{nasr2023scalable}                                    & extractable                        & verbatim                        & Pythia, LLaMA, InstructGPT, ChatGPT   & AuxDataset \cite{nasr2023scalable}                              & greedy                           &                   & 50                       \\
\citet{shao_quantifying_2023}                             & extractable                        & entity                         & GPT-Neo, GPT-J                          & Enron, LAMA \cite{petroni-etal-2019-language}                   & greedy                           &                   &                          \\
\citet{huang_are_2022}                                    & extractable                        & entity                         & GPT-Neo                                  & Enron                                               & greedy                           &                   &                          \\
\citet{yu2023bag}                                           & discoverable                       & verbatim                        & GPT-Neo-1.3B                             & lm-extraction-benchmark                             & greedy, top-p, top-k, nucleus & 50                & 100                      \\
\citet{biderman2023emergent}                                & discoverable                       & verbatim                        & Pythia                                   & PILE                                                & greedy                           & 32                & 64                       \\
\citet{ippolito2023preventing}                              & discoverable                       & approximate                     & GPT-3, PaLM                             & PILE                                                & greedy                           & 50                & 50                       \\
\citet{lee2022deduplicating}                                & extractable  discoverable       & verbatim                        & T5 [trained again]                       & C4 (variants)                      & top-k                            & \textgreater 50   &                          \\
\citet{k2022deduplicating}                                  & extractable                        & verbatim                        & Mistral project                          & OpenWebText, C4                                    & top-k, temprature               &                   & {100 - 700} (char) \\
\citet{lukas2023analyzing}                                  & extractable                        & entity                         & GPT-2                                    & ECHR, Enron                                        & greedy, beam                    &                   &                          \\
\citet{kim2023propile}                                      & extractable                        & entity                         & OPT                                      & PILE                                                & beam search                      &                   &                          \\
\citet{zhang2023ethicist}                                   & extractable                        & entity                         & GPT-Neo-1.3B                             & PILE                                                & greedy, top-p, top-k, beam    &                   &                          \\
\citet{ozdayi2023controlling}                               & discoverable                       & verbatim                        & GPT-Neo                  & lm-extraction-benchmark                             & beam                             & 50                & 50                
\\
\hline
\end{tabular}\\
\caption{Summary and the spectrum of memorization used in different studies: (1) The table highlights that the Pythia and GPT-Neo families are the most commonly used models in memorization studies. This is a logical choice, as they are open-source and available at multiple training checkpoints, making them particularly suitable for analyzing memorization dynamics and contributing factors. Similarly, the PILE dataset is the predominant dataset in these studies, which aligns with the fact that Pythia and GPT-Neo models are primarily trained on it, ensuring consistency in experimental settings. (2) In terms of generation settings, studies largely focus on producing 50-token or 32-token sequences, with greedy decoding being the dominant approach. This preference for greedy decoding is expected, as it maximizes the likelihood of generating memorized sequences, making it an effective choice for probing memorization tendencies. (3) Regarding the spectrum of memorization, approximately 60\% of studies focus on discoverability, while the remainder explores extractability. Similarly, verbatim memorization is examined in 60\% of works, whereas the rest are split between approximate memorization and entity-level memorization.}
\label{tab:dynamics-summary}
\end{table*}
}

\begin{tcolorbox}[colback=white, title=Answer to RQ3.]
In summary, memorization in LLMs is driven by model size, training data properties, input strategies, tokenization, decoding methods, and fine-tuning approaches. Larger models memorize more, duplicated data increases recall, and certain prompts signify memorization. Training dynamics show memorization grows with epochs but can also fade, as models tend to forget earlier samples. Additionally, interpretability research reveals memorization is distributed across layers, with early layers encoding and later layers amplifying content.
An overview of the findings for each of the factors discussed in this section is provided in Table~\ref{tab:findings}.\end{tcolorbox}

\section{Mitigating memorization: strategies and techniques}
\label{sec:mitigation}
As discussed in earlier sections, memorization is influenced by a range of factors and it is sometimes necessary for the learning process \cite{feldman2020does, brown2021memorization}. However, in scenarios where memorization could lead to privacy concerns or security vulnerabilities, some methods could be employed to mitigate its impact. In these cases, specific strategies are utilized to limit this phenomenon, ensuring that potential risks related to data exposure or misuse are minimized. In this section, we provide a holistic overview of such methods, categorized in three groups: \textit{data-level, training-time, and post-training approaches}.

\color{black}
\subsection{Data-level approaches} 
As discussed in section \ref{sec:training-data}, duplicated training data is one of the major factors of memorization. \citet{lee2022deduplicating} run exact matching and approximate matching (MiniHash) de-duplication algorithms on the C4 \cite{c4}, RealNews \cite{realnews}, LM1B \cite{lm1b}, and Wiki40B \cite{wiki40b} datasets and show that these datasets contain up to 13.6\% near duplicates and up to 19.4\% exact duplicates.

To investigate the impact of data de-duplication on a language model's memorization, they train a 1.5B parameter GPT-2 \cite{gpt2} model from scratch on three different settings: C4-Original, C4-NearDup, and C4-ExactSubstr, each for two epochs. Then they evaluate the memorization in no-prompt and prompted settings and measure the 50-token exact match (Section \ref{sub:exact-match}). The no-prompt experiment generations show $10\times$ less memorization in de-duplicated trained models. On the other hand, in the prompted experiment, when the prompt comes from the duplicate examples, the model trained on C4-Original generates the true exact continuation over 40\% of the time. The other two models also generate the ground truth more often when the prompt is sampled from the duplicate examples, suggesting that more harsh de-duplication algorithms are needed to prevent memorization.

\color{black}
\subsection{Training-time approaches}

\subsubsection{Differential privacy} \label{sec:dp}
Differential privacy is a data privacy method that ensures the results of any analysis over a dataset reveals minimal information about any individual’s data, protecting against privacy breaches \cite{DP}. This method is adopted in some techniques in machine learning to protect individual data points by adding noise, minimizing the impact of any single data point on the model’s output. \emph{DP-SGD} (Differentially Private Stochastic Gradient Descent) is an adaptation of the standard \emph{SGD} algorithm, designed to fine-tune language models while maintaining privacy \cite{dpsgd}. \citet{carlini2019canaries} demonstrate that by adjusting the privacy budget parameter $\epsilon$ in DP-SGD training, the exposure of memorized data can be reduced to a level that makes it indistinguishable from any other data. However, this comes at the cost of reduced model utility and a slower training process.

To address these utility issues, some studies propose selective differential privacy approaches \cite{kerrigan2020differentially, li2022large, shi2022just, shi2022selective}. For instance, \citet{kerrigan2020differentially} propose training a non-private base model on a public dataset and then fine-tuning it on a private dataset using DP-SGD. This approach aims to balance privacy with model performance.

\citet{li2022large} show that with carefully chosen hyperparameters and downstream task objectives, fine-tuning pretrained language models with DP-SGD can yield strong performance on a variety of NLP tasks at privacy levels. Remarkably, some of their fine-tuned models even outperform non-private baselines and models trained under heuristic privacy approaches.

While adopting the DP method in data privacy is mathematically proven to protect individuals' privacy, it is crucial to select hyperparameters wisely when used in training LLMs as a technique; otherwise, the model may not withstand stronger privacy attacks, potentially compromising its effectiveness as shown in \citet{lukas2023analyzing}.

\subsubsection{Promoting Reasoning}
A growing line of work aims to disentangle and control the interplay between reasoning and memorization in LLMs, offering tools to reduce memorization-driven failures and benchmark inflation. \citet{salido2025othersgeneraltechniquedistinguish} introduce ``None of the Others (NOTO)'', a simple modification to multiple-choice evaluations where the correct answer is replaced by ``None of the others,'' forcing models to reason by elimination rather than rely on memorized answer patterns. The large accuracy drops observed under NOTO reveal strong dependence on memorization and contamination in standard benchmarks. Complementary mechanistic insights are provided by \citet{hong-etal-2025-reasoning}, who show that the shift between reasoning and memorization is mediated by a single linear direction in the model's residual stream. By identifying and manipulating this reasoning feature, they demonstrate causal control over whether a model behaves like a reasoner or retrieves memorized content, suggesting targeted intervention as a potential mitigation path. \citet{du-etal-2025-reason} find that memorization is not a distinct or isolated mechanism but emerges through the same pathways that support reasoning. Even when models memorize noisy labels, they still execute intermediate reasoning steps, indicating that memorization is distributed and intertwined with reasoning. This challenges assumptions that memorization can be cleanly isolated, and motivates mitigation strategies that focus on shaping reasoning pathways rather than simply suppressing memory lookup.
Finally, \citet{li-goyal-2025-memorization} propose Memory Conditioned Training (MCT) to integrate new knowledge into LLMs. While standard updates improve memorization, reasoning over updated facts remains difficult. MCT prepends ``memory tokens'' during training, helping models recall new facts while still performing reasoning, highlighting the challenge of aligning memorization and reasoning in dynamic knowledge settings.

Together, these studies highlight strategies that encourage reasoning while managing memorization, providing a multi-faceted approach to mitigating failures driven by memorization.
\subsubsection{Training intervention methods}
\citet{AlternatingTeaching2024} propose the alternating teaching method, a teacher-student framework where multiple teachers trained on disjoint datasets supervise a student model in an alternating fashion to reduce unintended memorization. This approach demonstrates superior privacy-preserving results on the LibriSpeech dataset \cite{librispeech} while maintaining minimal utility loss when sufficient training data is available. 
\citet{Goldfish2024} introduce the ``goldfish loss,'' a subtle modification to the next-token training objective where randomly sampled subsets of tokens are excluded from the loss computation. This prevents models from memorizing complete token sequences, significantly reducing extractable memorization while maintaining downstream performance. 
\citet{ghosal2025memorization} propose ``\textit{Memorization Sinks},'' a training-based method that isolates memorized content into dedicated neurons, preventing it from interfering with general language knowledge and reducing the risk of memorizing sensitive information.

\subsection{Post-training approaches}

\subsubsection{Unlearning methods}
As memorization could lead to privacy risks and copyright issues, unlearning methods could be necessary in some situations. Methods like knowledge unlearning aim to selectively remove specific information from trained models without retraining them from scratch. \citet{bourtoule2021machine} introduce the ``SISA'' framework for efficient machine unlearning, which divides the training data into shards (shards partition the data into disjoint segments) and trains sub-models that can be easily retrained if data needs to be removed. For LLMs specifically, \citet{chen-yang-2023-unlearn} introduce lightweight unlearning layers into transformers, allowing for selective data removal without full model retraining.  \citet{pawelczyk2024incontext} introduce ``In-Context Unlearning,'' which involves providing specific training instances with flipped labels and additional correctly labeled instances as inputs during inference, effectively removing the targeted information without updating model parameters. \citet{DeMem2024} propose ``DeMem,'' a novel unlearning approach leveraging a reinforcement learning feedback loop with proximal policy optimization to reduce memorization. By fine-tuning the model with a negative similarity score as a reward signal, the approach encourages the LLM to paraphrase and unlearn pre-training data while maintaining performance. 

Additionally, knowledge unlearning techniques have been categorized into parameter optimization, parameter merging, and in-context learning, each offering unique advantages in efficiently removing harmful or undesirable knowledge from LLMs \cite{Si2023Knowledge}. These methods not only enhance privacy and security but also ensure that the overall performance of the models remains intact, making them scalable and practical for real-world applications \cite{yao2024largelanguagemodelunlearning}. 
\subsubsection{Model editing}
\citet{ruzzetti-etal-2025-private} propose ``\textit{Private Memorization Editing (PME),}'' a technique that detects memorized PII in a trained LLM and edits the model's internal memory to remove that PII, thereby reducing the risk of privacy leakage while preserving the model’s general language capabilities. In a similar line of thought, \citet{ni-etal-2025-controllable} introduce a weight‑pruning based method that provides a unified framework to tune a model's memorization: by gradient‑guided pruning, one can suppress or amplify memorization depending on privacy vs utility needs, while balancing general language performance.


\color{black}

\subsubsection{Decoding-based methods}
\citet{ippolito2023preventing} propose ``MemFree Decoding,'' a novel sampling strategy to reduce memorization during text generation by avoiding the emission of token sequences matching the training data. They demonstrate its effectiveness across multiple models, including GPT-Neo \cite{gpt-neo} and Copilot \cite{copilot}, significantly lowering generation similarity with the training data. However, they also reveal that a simple style transfer in prompts can bypass this defense, highlighting its limitations. Similarly, \citet{NucleusMem2024} investigate the impact of nucleus sampling on memorization, showing that while increasing the nucleus size slightly reduces memorization, it only provides modest protection. They also highlight the distinction between ``hard'' memorization, involving verbatim reproduction, and ``soft'' memorization measured by the ROUGE similarity metric \cite{rouge}, where generated outputs echo the training data without exact replication. 

\begin{remark}
Data-level mitigations can reduce memorization but have limitations: deduplication may hurt task performance, and data sanitization is not fully reliable \cite{lukas2023analyzing, pham2025largelanguagemodelsreally}. Training interventions, including DP, alternating teaching, and other heuristic approaches, offer stronger control during training. DP provides formal privacy guarantees but depends on hyperparameters and can still leave residual memorization. Other methods may interfere with long-range dependencies or internal representations, affecting downstream reasoning.

Post-training methods, such as unlearning, model editing, and decoding-based strategies, allow targeted mitigation without retraining, but they rely on identifying memorized content, risk unintended side effects, and cannot fully address hidden memorization or reasoning influences.

Together, these approaches highlight that while mitigation is possible, no single method fully eliminates memorization, and careful combination is needed to balance privacy, reasoning, and performance.

\end{remark}

\begin{tcolorbox}[colback=white, title=Answer to RQ4.]
Data deduplication has been widely shown to be one of the most effective and efficient methods to mitigate memorization. Differential privacy, unlearning, and other heuristic approaches can also be used to prevent memorization but those need to be employed precautiously as they might hurt the model performance.
\end{tcolorbox}

\section{Future directions and open challenges}
\label{sec:future}

Based on the discussion of the existing literature to date on memorization in LLMs, we make suggestions for research topics to be addressed in the near future.

\subsection{Balancing memorization benefits and privacy risks}
As discussed earlier, memorization in LLMs can have beneficial uses, such as improving factual recall and enhancing performance in specific tasks like question answering. At the same time, privacy and copyright concerns remain major challenges \cite{neel2023privacy, smith2023identifying, yao2024survey}. Interestingly, recent work challenges the belief that mere verbatim memorization leads to privacy leakage; concluding that what drives sensitive information exposure is how memorized content is utilized or revealed by the model \cite{sander2025rethinking}. On the other hand, standard privacy-enhancing technologies, such as multi-party computation, homomorphic encryption, and differential privacy, can mitigate memorization risks, but often reduce model performance by limiting precise retention or perturbing outputs. Future research should focus on strategies that balance memorization benefits with privacy protection, developing techniques that safeguard sensitive data and intellectual property without significantly degrading accuracy or utility. Such approaches are essential for ensuring both high-performance outcomes and legal compliance in LLM deployment.

\subsubsection{Memorization and Differential Privacy}
As discussed in Section \ref{sec:counterfactual}, counterfactual memorization shares some conceptual similarities with differential privacy. Counterfactual memorization evaluates the difference in a model's behavior on a specific data point when the data point is included in or excluded from the training dataset. In contrast, a model is considered differentially private for a given example if the outputs of two models trained on neighboring datasets—differing by only that example—are indistinguishable.

In the context of text data, the definition of a ``data point'' can vary significantly. It might refer to named entities, individual sentences, complete documents, or even the entire text corpora. Both of these points underscore the need for further research to examine the interplay and dynamics between memorization and differential privacy at different data granularities.

\subsection{Reducing verbatim memorization in favor of content memorization}
The interplay between factual and content memorization is relevant in LLM development, because there is a trade-off between the correctness of information, and undesirable recall of training data: Although verbatim memorization can provide more precise information recall, it also presents higher risks in terms of privacy and data protection. Content memorization, on the other hand, contributes to the model's ability to generalize and apply knowledge flexibly, but it may also present more challenges for audit and control. For example, when using LLMs for question answering, literal reproduction of facts is desired; this includes names and numbers (e.g., ``Who was the first person to fly across the ocean and when did this take place?''). It is straightforward to evaluate LLMs to generate the (correct) facts. If content memorization is preferred over verbatim memorization in LLMs, a higher rate of hallucinations should be accepted because the model generates text more freely based on its parametric knowledge.

Future research in this area should therefore focus on developing techniques to balance these two types of memorization, enhancing the benefits of each while mitigating their undesired consequences. This could involve methods to selectively encourage content memorization while limiting unnecessary factual memorization, particularly of sensitive information, depending on the type of task at hand.

\subsection{The boundary between memorization and understanding}
It is relatively straightforward to design experiments that demonstrate a model's ability to generate fluent and human-like text, such as generating novel content or adapting to new contexts. However, verifying that these capabilities are not simply the result of memorization rather than the abstraction/generalization over the information processed during training is much more challenging, as both can produce similar outcomes. Distinguishing between these requires carefully designed experiments, and further research is needed to clarify when a model makes abstractions over learned concepts versus merely memorizing and reproducing them.

\color{black}
\subsection{Memorization in specific contexts}
Several application domains are currently understudied with respect to the effect of memorization. We identified four contexts specifically where more research is needed. 

\paragraph{Conversational Agents} LLMs that have been fine-tuned for conversations can be used as conversational agents (chatbots), e.g., to assist customers of online services~\cite{xi2023rise,deng2023rethinking}. \citet{nasr2023scalable} introduce their so-called ``divergence attack'' to extract memorized samples from conversation-aligned LLMs. As they also discuss, these attacks are not powerful enough to stimulate training data reproduction. However, this does not mean that the conversation-aligned LLMs are not vulnerable to attacks or manipulation. It also means This makes us conclude that attacking conversation-aligned language models requires more advanced methods. Since most of the production language models are only available in a conversational settings, addressing memorization in conversational agents and conducting novel attacking methods to alleviate training data extraction in these models would be a prominent research direction.

\paragraph{Retrieval-Augmented Generation (RAG)} In RAG frameworks, LLMs are helped with a retrieval component that first selects the most relevant documents from a collection and feeds them to the prompt \cite{lewis2020retrieval,gao2023retrieval}. The LLM then generates an answer based on the provided sources. This has advantages such as the reduction of hallucination \cite{shuster2021retrieval}, the transparency of sources, and the potential of generating answers related to proprietary or novel sources that were not in the LLM training data. Although very popular, RAG is not yet thoroughly analyzed regarding memorization. This is important, because a reduction of hallucination when using RAG in LLMs could also have an increase in memorization as a side effect: less hallucination means that the LLM stays closer to the original content. Recently, \citet{zeng2024goodbadexploringprivacy} analyzed privacy aspects in RAG. They show that RAG systems are vulnerable to leaking private information. On the other hand, they also found that RAG can mitigate the reproduction of the LLM training data. A broader study focusing on memorization in RAG-based LLMs that compares different proposed retrieval architectures is an important future research direction.

\paragraph{Multilingual Large Language Models (xLLM)} xLLMs are trained to interpret and generate text across multiple languages, often relying on large multilingual corpora to develop language-agnostic representations \cite{qin2024multilingual}. Yet, they commonly suffer from data scarcity in low-resource languages, resulting in notable performance disparities \cite{xu2024survey}. Recent work highlights that memorization in multilingual models is not uniform across languages. \citet{luo-etal-2025-shared} show that memorization patterns are shaped by cross-lingual similarities rather than data volume alone, and \citet{satvaty-etal-2025-memorization} demonstrate that lower-resource languages show different memorization patterns when compared to the English language. These findings motivate future work to more systematically analyze how memorization and privacy risks differ across languages and to uncover the underlying factors driving these differences, and to develop methods that effectively mitigate them.

\paragraph{Diffusion Language Models (DLM)} Since DLMs show remarkable performance in the vision domain, researchers have started to adopt the diffusion models (DM) idea to the text domain and utilize their generative capabilities \cite{zou2023survey}. \citet{carlini2023extracting} have conducted data extraction analysis on the image diffusion models, showing  that diffusion models are much less private than prior generative models such as GANs, and that mitigating these vulnerabilities may require new advances in privacy-preserving training. \citet{gu2024on} also show that according to the training objective of the diffusion models, a memorization behavior is theoretically expected and then they quantify the impact of the influential factors on the memorization behaviors in DMs. However, research focusing on the memorization issues related to DLMs for text remains unexplored. Even though the idea of diffusion language models is the same as the vision domain, they are inherently different, because of the discrete nature of the text domain. Therefore, an analysis of the general vision diffusion models may not be applicable to the diffusion language models, making an independent research on memorization against DLMs a prominent future direction.

\begin{tcolorbox}[colback=white, title=Answer to RQ5.]
Key aspects of memorization in LLMs still require exploration: Balancing privacy with performance remains challenging, as does reducing verbatim recall while preserving factual accuracy. Distinguishing memorization from true understanding needs better evaluation methods. The interplay between memorization and differential privacy at different granularities is underexplored. Additionally, specific contexts like conversational agents, RAG systems, multilingual models, and diffusion language models require deeper study to understand their unique memorization risks.
\end{tcolorbox}

\section{Conclusion}
\label{sec:conclusion}
In this paper, we organized, summarized, and discussed the existing scientific work related to undesirable memorization in LLMs. Undesirable memorization might lead to privacy risks and other ethical consequences.
We found that there exists a large body of research on  memorization in LLMs, given the young age of the technology: Transformer models were first developed in 2017~\cite{attention}, and the first generative large language model with emergent abilities was released in 2022~\cite{gpt3,wei2022emergent}. The majority of papers on the topic were therefore published in recent years. 
This indicates that the field is working fast to analyze memorization and developing methods to mitigate undesirable memorization.

Despite the fast-growing body of literature on the topic, we argue that there are important areas that require more attention in research in the coming years. We have pointed out four specific contexts in which memorization needs to be studied and, when needed and possible, mitigated: LLM-based conversational agents, retrieval-augmented generation, multilingual LLMs, and diffusion language models. 
These areas all are of high importance, not only from the academic perspective, but maybe even more from the application and industry perspective. In particular, conversational agents and retrieval-augmented generation are actively being developed in the commercial context. When applied to proprietary databases or in interaction with customer information, these applications are particularly vulnerable to privacy and security risks.

\section*{Limitations}
The main limitation of a survey paper on a highly active research topic is the fast pace at which the field is evolving. A survey paper written in 2025 risks becoming outdated by 2026. Although we acknowledge this, we argue that the topic of memorization is too important to overlook, especially given the added value we provide through our concrete suggestions for future directions. To mitigate this issue, we maintain a dedicated GitHub repository\footnote{\url{https://github.com/alistvt/undesirable-llm-memorization}} that catalogs the references discussed in this survey and will be regularly updated to reflect the latest developments in the field.

\section*{Acknowledgment}
This publication is part of the project LESSEN\footnote{\url{https://lessen-project.nl}} with project number NWA.1389.20.183 of the research program NWA ORC 2020/21 which is (partly) financed by the Dutch Research Council (NWO).

\color{black}
\bibliographystyle{IEEEtranN}

\bibliography{references}

\end{document}